%% file: graspflow.tex
\documentclass[letterpaper, 10 pt, conference]{ieeeconf}

\IEEEoverridecommandlockouts

\usepackage{cite}

\input{content/math_commands.tex}

\usepackage{amsmath,amssymb,amsfonts}
\usepackage{graphicx}
\usepackage{textcomp}
\usepackage{color}
\usepackage{xcolor}

\usepackage{hyperref}       
\usepackage{booktabs}       
\usepackage{nicefrac}       
\usepackage{microtype}      
\usepackage{mathrsfs}
\usepackage{comment,multirow,graphicx,bbm,subcaption,caption,mathtools}
\usepackage{mathabx,algorithmic,algorithm}

\usepackage{balance}
\usepackage[normalem]{ulem}  

\graphicspath{ {figs/} }

\newcommand{\para}[1]{\vspace{0.4em}\noindent\textbf{#1}}

\newcommand{\method}{\textsc{GraspFlow}}
\newcommand{\graspflow}{\textsc{GraspFlow}}
\newcommand{\criteria}{\mathbf{c}}
\newcommand{\grasp}{\mathbf{g}}
\newcommand{\obs}{\mathbf{o}}
\newcommand{\btheta}{\boldsymbol{\theta}}

\newcommand{\rev}[1]{{#1}}

\usepackage{mathabx,algorithmic,algorithm}

\title{\LARGE \bf
Refining 6-DoF Grasps with Context-Specific Classifiers
}

\author{Tasbolat Taunyazov$^{1}$, Heng Zhang$^{1}$, John Patrick Eala$^{1}$, Na Zhao$^{2}$, and Harold Soh$^{1,3}$
\thanks{This research is supported by the National Research Foundation, Singapore under its Medium Sized Center for Advanced Robotics Technology Innovation.}
\thanks{$^{1}$Authors are with the Dept. of Computer Science, National University of Singapore. Correspondence to Tasbolat Taunyazov 
        {\tt\ tasbolat@comp.nus.edu.sg} or Harold Soh {\tt\ harold@comp.nus.edu.sg}}
\thanks{$^{2}$Author is with Singapore University of Technology and Design. Email: 
        {\tt\ \url{na_zhao@sutd.edu.sg}}}
\thanks{$^{3}$Smart Systems Institute, National University of Singapore}
}

\begin{document}

\maketitle

\input{content/abstract}

\IEEEpeerreviewmaketitle

\thispagestyle{empty}
\pagestyle{empty}

\input{content/introduction}
\input{content/method}
\input{content/instantiation}
\input{content/experiments}

\input{content/discussion}

\balance

\bibliography{graspflow}
\bibliographystyle{IEEEtran}

\end{document}

%% file: content/math_commands.tex
\usepackage{amsmath,amsfonts,bm}

\def\eqref#1{equation~\ref{#1}}

\def\1{\bm{1}}

\def\rvg{{\mathbf{g}}}

\def\rvw{{\mathbf{w}}}

\DeclareMathAlphabet{\mathsfit}{\encodingdefault}{\sfdefault}{m}{sl}
\SetMathAlphabet{\mathsfit}{bold}{\encodingdefault}{\sfdefault}{bx}{n}

\def\gF{{\mathcal{F}}}

\def\gH{{\mathcal{H}}}

\def\gN{{\mathcal{N}}}

\def\gP{{\mathcal{P}}}

\def\gW{{\mathcal{W}}}
\def\gX{{\mathcal{X}}}

\def\sR{{\mathbb{R}}}

%% file: content/abstract.tex
\begin{abstract}
In this work, we present GraspFlow, a refinement approach for generating context-specific grasps. We formulate the problem of grasp synthesis as a sampling problem: we seek to sample  from a context-conditioned probability distribution of successful grasps. However, this target distribution is unknown. As a solution, we devise a discriminator gradient-flow method to evolve grasps obtained from a simpler distribution in a manner that mimics sampling from the desired target distribution. Unlike existing approaches, GraspFlow is modular, allowing grasps that satisfy multiple criteria to be obtained simply by incorporating the relevant discriminators. It is also simple to implement, requiring minimal code given existing auto-differentiation libraries and suitable discriminators. Experiments show that GraspFlow generates stable and executable grasps on a real-world Panda robot for a diverse range of objects. In particular, in 60 trials on 20 different household objects, the first attempted grasp was successful 94\% of the time, and 100\% grasp success was achieved by the second grasp. Moreover, incorporating a functional discriminator for robot-human handover improved the functional aspect of the grasp by up to 33\%. 
\end{abstract}

%% file: content/introduction.tex
\section{Introduction} 

The right way to grasp an object is context-specific. It depends not only on the object being grasped, but also on the characteristics and state of the robot executing the grasp --- certain grasp configurations (and subsequent lifting/manipulation) are achievable on some robots, yet not on others. In addition, grasps should be functional; a robot grasp that works well for a pick-and-place operation may be inappropriate for human-robot handover. How robots can synthesize grasps that satisfy such diverse context-dependent quality criteria remains a fundamental challenge in robotics. 

\begin{figure}
\centering
	\includegraphics[width=0.8\columnwidth]{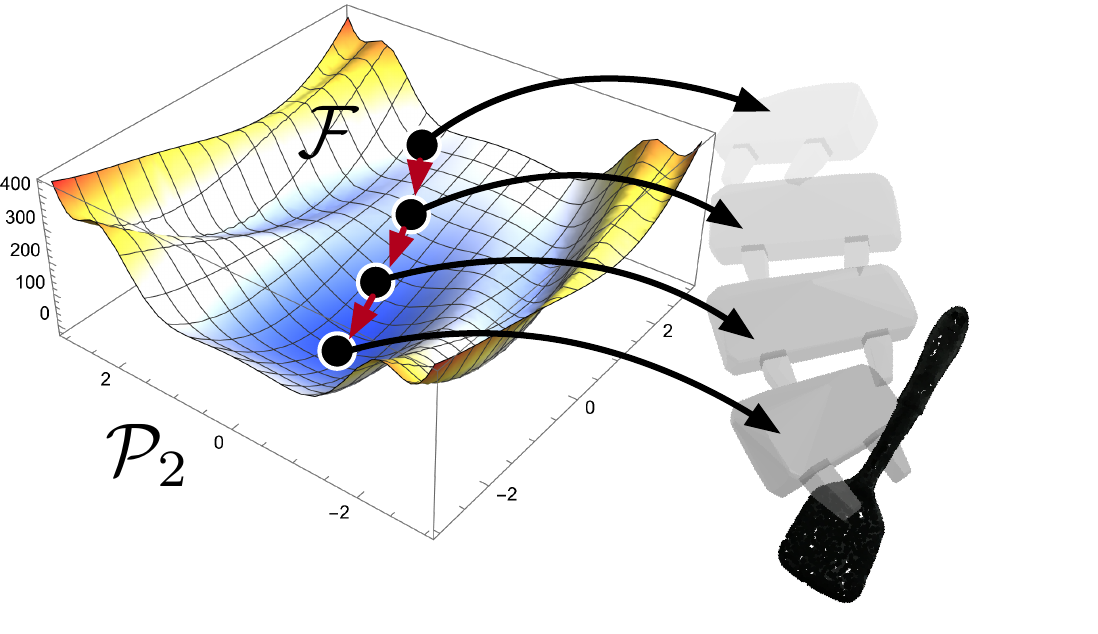}
	\caption{\small Illustration of \graspflow{}. Grasps shown are samples from probability distributions (black circles) along the gradient flow from the initial distribution $q_0$ to a target distribution $p$.}
	\label{fig:mainfig}
\end{figure}

In this work, we adopt a new perspective on the problem of grasp synthesis and treat the problem as one of sampling: we seek to sample quality grasps from a context-dependent target distribution $p$. The key problem is that we don't have access to $p$ and it is unknown. We propose a method that \emph{evolves}, or \emph{refines}, grasps generated from a simpler base distribution $q_0$ (e.g., a deep generative model~\cite{mousavian20196}) such that the grasps appear to be sampled from $p$. Our approach is based on theory of discriminator gradient flows~\cite{fatir2021refining}; we leverage a Fokker-Planck equation that represents the gradient flow --- the steepest descent curve in the space of probability measures --- obtained from minimizing a regularized $f$-divergence functional between $q_0$ and $p$ (Fig. \ref{fig:mainfig}). 

In this setup, the density ratio $q_0/p$ emerges as the crucial quantity to be estimated. We propose an approximation using discriminators/classifiers trained (or designed) to evaluate grasp quality criteria. This enables a modular approach towards grasp synthesis; different discriminators can be applied to suit specific contexts. Moreover, while classifiers have become strongly associated with deep learning, our method does not specifically rely on neural methods. Practitioners are free to select classifiers which are best for the task (even handcrafted models), as long as they are differentiable. Our approach, which we call \graspflow{}, is straightforward to implement, requiring only a few lines of code given modern auto-differentiation libraries and existing discriminators. 

Experiments show that \graspflow{} is able to generate stable and executable 6DOF grasps on a Franka-Emika Panda robot. In our first experiment, \graspflow{} significantly improves grasps sampled from the GraspNet VAE~\cite{mousavian20196} on a set of 20 different household objects (1800 grasps); success percentages rose from 42\% to 82\% after sample refinement. Note that this was achieved \emph{without} having to discard grasps that violate robot kinematics as in prior work (e.g., \cite{mousavian20196}). In a second experiment involving 120 grasps, we demonstrate that successful \emph{and} functional grasps can be obtained when a simple handover discriminator was incorporated into the setup; on two objects (a hammer and spatula), the proportion of functionally-appropriate grasping increased from 40\% to 68\%. In both experiments, a successful grasp was obtained within the first \emph{two} attempts.  


\para{Relation to Prior Work.} \graspflow{} pertains to the study of robot grasping, which has a rich history and encompasses a variety of approaches~\cite{bohg2013data,shimoga1996robot,sahbani2012overview,kleeberger2020survey}. Recent work has focused largely on data-driven methods using deep learning to generate suitable grasps from low-level observations (e.g., RGB-D or point clouds) and derive models that generalize to novel  objects~\cite{lenz2015deep,ten2017grasp,mahler2017dex,levine2018learning,satish2019policy,morrison2018closing}. \rev{For example, DexNet 2.0~\cite{mahler2017dex} employs a trained convolutional neural network to rank and filter potential planar grasps.}
The closest related body of work to ours involve methods that synthesize (or refine) grasps via optimization of learnt functions. Compared to end-to-end deep models, these methods separate out the optimization/inference process from grasp evaluation. As such, they can generate a diverse set of grasps for a given object and filter out those that do not meet selected criteria; in contrast, single-stage end-to-end models typically have to be completely retrained for new quality criteria. Early work by \cite{zhou20176dof} optimized a 6DOF grasp quality function via gradient descent and quasi-Newton methods. Later methods~\cite{lu2020planning,lu2020multifingered,van2020learning} viewed grasp synthesis/planning as probabilistic inference --- \cite{lu2020planning} proposed to conduct maximum likelihood estimation of the grasp by optimizing the likelihood of grasp success conditioned upon the grasp pose. This probabilistic approach was later extended to incorporate priors, yielding a maximum a posteriori solution~\cite{lu2020multifingered,van2020learning}. GraspNet~\cite{mousavian20196} refines grasps sampled from a variational autoencoder (VAE) by optimizing a quality function (similar in spirit to \cite{zhou20176dof}), while later variants~\cite{murali20206,yang2021reactive} performed Monte-Carlo sampling using the discriminator to compute the acceptance ratio. 

\graspflow{} can be seen as unifying the sampling, probabilistic, and optimization approaches towards grasp synthesis. As such, it inherits many of the advantages above, yet resolves issues associated with each individual approach. \graspflow{}'s basic formulation is one of sampling, which enables the incorporation of uncertainty and provides a diversity of grasp candidates in proportion to their estimated probability of success --- this can be difficult to achieve in a pure optimization framework. However, a standard sampling-based approach can be wasteful since many potential grasps are filtered away. Rather than discard samples, \graspflow{} \emph{refines} samples via a scheme derived by optimizing the sampling distribution; this method enables us to sample from a desired target distribution of successful grasps, which remains implicit in the setup. To our knowledge, \graspflow{} is the first work to optimize for multiple criteria in such a manner.

%% file: content/method.tex
\section{\method{}: Grasp Refinement via Discriminator Gradient Flow}

In this section, we describe \method, a framework for generating grasp candidates that satisfy desired criteria. At a high-level, we seek to sample grasps $\grasp$ from a conditional distribution $p(\grasp|\criteria, \obs)$ where $\criteria$ and $\obs$ are the desired quality criteria and robot observation, respectively. Unfortunately, $p$ is generally unknown. Instead, we only have access to a distribution $q_0(\grasp|\criteria, \obs)$ that is easy to sample from. We will develop a gradient-flow formulation that will enable us to \emph{transform} grasps sampled from $q_0(\grasp|\criteria, \obs)$ such that they appear to be coming from $p(\grasp|\criteria, \obs)$. We will begin with a brief introduction to gradient flows\footnote{Please see \cite{santambrogio2017euclidean} for a more thorough introduction.}, followed by a description of how gradient flows can be applied to grasp refinement using multiple classifiers, which enables sampling for a specified context. In this section, we will focus on conveying the main ideas behind our approach and delay discussing implementation details (e.g., grasp representation, the specific classifiers used in our setup) to the next section. \rev{Our presentation summarizes Ansari et al.~\cite{fatir2021refining}, with additional comments regarding key differences when applying discriminator gradient flows to grasping.}

\para{Background on Gradient Flows.}
To provide intuition, we begin with the familiar notion of a Euclidean space with the 2-norm $(\gX,\|\cdot\|_2)$. Given a smooth energy function $F: \gX \rightarrow \sR$, the curve $\{\rvg_t\}_{t\in\sR_+}$ that follows the direction of steepest descent and minimizes the energy is called the \emph{gradient flow}, $\rvg'(t) = -\nabla F(\rvg(t))$. 
In this work, we are interested in sampling from a \emph{probability distribution} of successful grasps. Rather than Euclidean spaces, we are interested in steepest descent curves in the metric space of \emph{probability measures}. Specifically, we will examine gradient flows in the 2-Wasserstein space ($\gP_2(\Omega), \gW_2$), i.e., the space of probability measures with finite second moments $\gP_2(\Omega)$ that is coupled with the Wasserstein metric $\gW_p$. Given a functional $\gF: \gP_2(\Omega) \rightarrow \sR$ in the 2-Wasserstein space, the gradient flow $\{q_t\}_{t\in\sR_+}$ of $\gF$ minimizes the value of $\gF$. 
Recently, gradient flows been used in deep generative modeling~\cite{fatir2021refining,liutkus2019sliced,gao2019deep,gao2020learning}. We adopt a similar scheme to \cite{fatir2021refining}, who introduced a gradient flow-based technique for refining samples (images and text) using a GAN-based discriminator. Here, we will refine robot grasps, which unlike the samples in \cite{fatir2021refining}, are conditionally-generated and have to be refined to satisfy  quality criteria specific to robot grasping.

\para{Grasp Refinement via Gradient Flow.}
\rev{We adopt a standard definition of a grasp, i.e., a set of contact points with an object $j$ which restricts movement when external forces are applied~\cite{bicchi2000robotic}. We model $p(\grasp|j)$ as a uniform distribution over end-effector poses $\grasp$ that form grasps. In practice, two issues arise: (i) we do not know the object $j$ but have access to an observation $\obs$, and (ii) grasps may not satisfy desired criteria $\criteria$ (e.g., stable, functional). Hence, we seek to sample from the context-dependent distribution 
$p(\grasp|\criteria,\obs) \propto p(\criteria|\grasp,\obs)p(\grasp|\obs)$
where we have applied Bayes rule and marginalized out the unobserved object, $p(\grasp|\obs) = \sum_j p(\grasp|j)p(j|\obs)$. The distribution $p(\grasp|\criteria, \obs)$ is generally difficult (or impossible) to explicitly specify or compute.}

We only have access to a distribution $q_0(\grasp|\criteria, \obs)$ that is easy to sample from, but may not generate grasps that fulfill the desired criteria. We refer to the  variables $(\criteria, \obs)$ as the \emph{context}. 
Depending on the chosen distribution, $q_0$ may disregard the context or parts of it. To reduce clutter, we will drop the explicit dependence on $\criteria$ and $\obs$ but note that $p$ is always conditioned upon the context. 

Our goal is to obtain grasp candidates from $p$ rather than $q_0$. To achieve this, we first consider how we can transform $q_0$ into $p$. Similar to how we might optimize samples in a Euclidean space by minimizing a function, we minimize the entropy-regularized $f$-divergence functional:
\begin{align}
	\gF_{p}^f(q) 
	&\triangleq \underset{f\text{-divergence } \mathbb{D}_f[p(\rvg)\|q_0(\rvg)]}{\underbrace{\int f\left(q_0(\rvg)/p(\rvg)\right)p(\rvg)d\rvg}} +\gamma\underset{\text{negative entropy } -\gH(q_0(\rvg))}{\underbrace{\int q_0(\rvg) \log q_0(\rvg)d\rvg}},
\end{align}
where the $f$-divergence term $\mathbb{D}_f[p\|q_0]$ captures the ``distance'' between a density $q_0$ and our target $p$. 
The gradient flow of $\gF_p^f(q)$ in the Wasserstein space is given by the  Fokker-Planck equation (FPE),
\begin{align}
	\partial_tq_t(\rvg) = \nabla_\rvg\cdot\left(q_t(\rvg)\nabla_\rvg f'\left(q_t(\rvg)/p(\rvg)\right)\right) + \gamma\Delta_{\rvg\rvg}q_t(\rvg)  \label{eq:fokker-plankpde}
\end{align}
where $f'$ is the derivative of the chosen $f$-divergence, and $\nabla_\rvg\cdot$ and $\Delta_{\rvg\rvg}$ denote the divergence and the Laplace operators, respectively.  Eq. (\ref{eq:fokker-plankpde}) gives us the gradient flow of the density $q_t$ as it is transformed from $q_0$ to $p$, but working directly with the distributions $q_t$ is difficult. Ideally, we would like to transform \emph{samples}, i.e., the candidate grasps. 

\para{From Distributions to Samples.} The FPE above has an equivalent Stochastic Differential Equation (SDE) formulation~\cite{risken1996fokker}:
\begin{align}
	d\rvg_t = \underset{\text{drift}}{\underbrace{-\nabla_\rvg f'\left(q_t(\rvg_t)/p(\rvg_t)\right)dt}}+\underset{\text{diffusion}}{\underbrace{\sqrt{2\gamma}d\rvw_t}},
	\label{eq:mckean-vlasov-process}
\end{align}
which describes the evolution of \emph{samples} following the distributions in Eq. (\ref{eq:fokker-plankpde}). This SDE 

can be numerically solved using the Euler-Maruyama method~\cite{beyn2011numerical}:

\begin{align}
	\rvg_{\tau_{n+1}} = \rvg_{\tau_n} -\eta\nabla_\rvg f'\left(q_{\tau_n}(\rvg_{\tau_n})/p(\rvg_{\tau_n})\right) + \sqrt{2\gamma\eta}\bm{\xi}_{\tau_n},
	\label{eq:euler-maruyama}
\end{align}
where $\bm{\xi}_{\tau_n} \sim \gN(\bm{0}, \mathbf{I})$. The $\tau_n$ are discretized time-steps where we have partitioned the time interval $[0, N]$ into equal intervals of size $\eta$. Unlike the FPE, Eq. (\ref{eq:euler-maruyama}) provides a method for changing grasp candidates drawn from $q_0$ to grasps from $p$.

\para{Approximating the Density Ratio with Context Classifiers.} To transform the samples, Eq. (\ref{eq:euler-maruyama})  requires evaluation of the density-ratio  $q_{\tau_n}(\rvg_{\tau_n})/p(\rvg_{\tau_n})$. \cite{fatir2021refining} approximated this density ratio using the \emph{density-ratio trick}~\cite{sugiyama2012density}, i.e., a differentiable discriminator trained to distinguish between samples from $q_0$ and samples from $p$, 
\begin{align}
	\frac{q_{\tau_n}(\rvg_{\tau_n})}{p(\rvg_{\tau_n})} \approx \frac{1- p(d=1|\rvg_{\tau_n})}{p(d=1|\rvg_{\tau_n})}
\end{align}
where $d=1$ is the label given to samples from $p$ (and $d=0$ for samples from $q_0$). This type of classifier may be familiar to readers acquainted with GANs; similar classifiers are learned during GAN training to tell apart the real and generated samples. 
In our setting however, we do \emph{not} have ready access to a classifier of this nature. Indeed, gathering a large number of successful real-world grasps to train such a discriminator for a variety of contexts would be prohibitively expensive. 

Instead, we propose a different approximation using a set of classifiers developed for assessing grasp quality. Specifically, we approximate the density ratio, 
\begin{align}
	\frac{q_{\tau_n}(\rvg_{\tau_n})}{p(\rvg_{\tau_n})} \approx \frac{1 - \prod_i p(c_i = 1 | \rvg_{\tau_n}, \obs)}{\prod_i p(c_i = 1 | \rvg_{\tau_n}, \obs)} \label{eq:multipleclass}
\end{align}  
This form assumes that the different quality criteria are conditionally independent \emph{given the grasp and observation}, which is not unreasonable if $\grasp$ and $\obs$ already comprise the information necessary to determine whether a given criterion $c_i$ is fulfilled. 
Also, the classifiers $p(c_i | \grasp, \obs)$ are not trained to distinguish samples from $q$ and $p$; an implicit assumption is that $q$ is similar to the distribution of grasps used for training the classifiers and that do not fulfill all the quality criteria. 
If this property does not hold for a specific sampler $q$, we can amend the flow using a corrector term. 
We did not find it necessary to employ this corrector in our experiments, presumably because the grasps generated from many samplers, even state-of-the-art deep generative models, are unlikely to fulfill all desired criteria.

To obtain our final refinement process, we combine  (\ref{eq:mckean-vlasov-process}) and (\ref{eq:multipleclass}) to give,
\begin{align}
	\rvg_{\tau_{n+1}} = \rvg_{\tau_n} &
   - \eta\nabla_{\rvg_{\tau_n}} f'\left(\frac{1 - \prod_i p(c_i = 1 | \rvg_{\tau_n}, \obs)}{\prod_i p(c_i = 1 | \rvg_{\tau_n}, \obs)}\right) \nonumber \\ & + \sqrt{2\gamma\eta}\bm{\xi}_{\tau_n},
	\label{eq:graspflow}
\end{align}
The sample evolution equation above is simple to implement given auto-differentiation libraries and affords a degree of modularity---different classifiers can be selected and combined to jointly refine grasps for a particular application 

\para{Summary and Practical Use.} In the \method{} framework, we first choose an base sampler $q_0$ along with classifiers $p(c_i|\grasp,\obs)$ representing our desired quality criteria. We also select a $f$-divergence; in our experience, the KL-divergence $f(r)=\log(r)$ ($f'(r) = \log(r)+1)$) generally works well, but other divergences can be applied. We then sample a candidate grasp $\rvg_{\tau_0} \sim q_0$, and then update the grasp using Eq. (\ref{eq:graspflow}) for $N$ time steps. The hyperparameters $\eta$, $N$, and $\gamma$ can be tuned for performance. Given that $\eta$ emerges as part of a numerical solution for the SDE (\ref{eq:mckean-vlasov-process}), it is generally set to be small (e.g., $10^{-3}-10^{-5}$). $N$ should be chosen based on the sampler $q_0$; larger $N$ is needed for samplers that are generate poor grasps (i.e., $q_0$ is far from $p$). Finally, $\gamma$ can be set to  $10^{-2}-10^{-4}$ to allow for diversity in the generation. In our experiments, we found \method{} to perform similarly within reasonable parameter ranges.

%% file: content/instantiation.tex
\section{\method{} for Stable, Executable, and Functional 6DOF Grasps}
\label{sec:graspflow6dof}

\begin{figure*}
    \centering
    \includegraphics[width=1.0\linewidth]{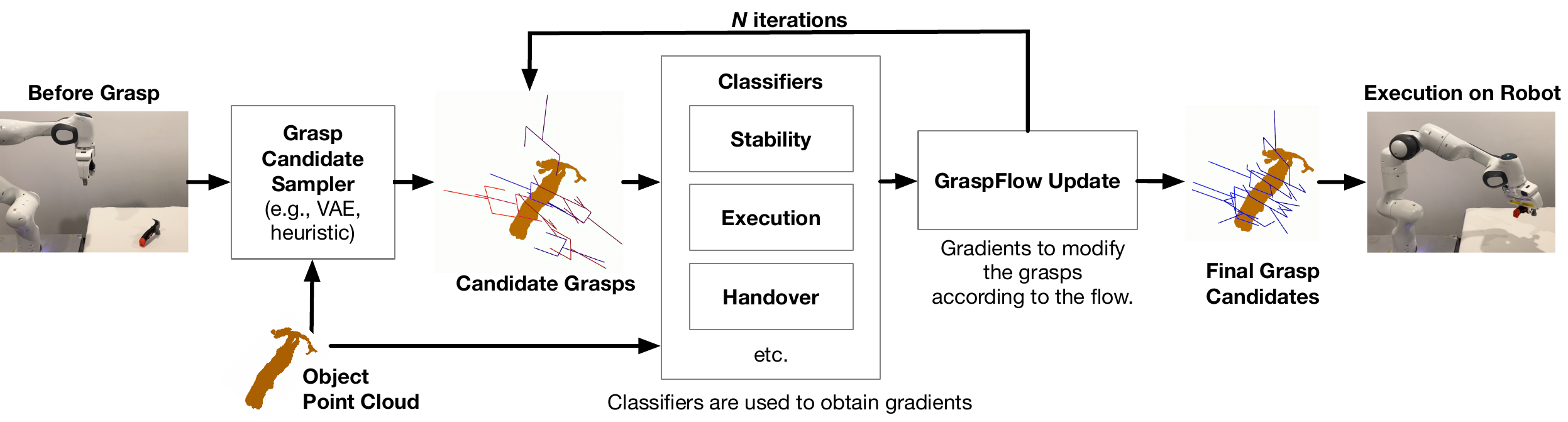}
    \caption{\rev{\small Overview of \method{} for 6DOF grasping. \method{} refines sampled grasp candidates over $N$ iterations using gradients obtained from the classifiers. Here, a batch of grasps are refined and candidates can be ranked using the classifiers before execution.}}
    \label{fig:graspflow_overview}
\end{figure*}

Thus far, we have described \method{} in a relatively abstract manner. Here, we will describe how \method{} can be used to obtain desired 6-DoF grasps for a Franka-Emika Panda arm \rev{(Fig. \ref{fig:graspflow_overview})}. Each hand pose or grasp $\grasp$ is represented by its rotation and translation $(\mathbf{r}, \mathbf{t}) \in SE(3)$ where $\mathbf{r} \in SO(3)$ and $\mathbf{t} \in \mathbb{R}^3$. We mainly focus on obtaining stable and executable grasps using (i) a stability classifier trained in simulation to distinguish stable from unstable grasps, and (ii) a handcrafted ``execution'' classifier that estimates whether the grasp can be performed by the Panda. Finally, we perform preliminary experiments with (iii) a functional classifier that classifies grasps for robot-human handover. Alternative classifiers can be used without changing the overall framework. 

\subsection{Stability Classifier}

Our data-driven grasp stability classifier is based on the GraspNet  evaluator~\cite{mousavian20196}, which uses  PointNet~\cite{qi2017pointnet}. The input to the classifier is a point cloud comprising the object point cloud (obtained from 4 views) and the gripper point cloud, with a feature label to distinguish the two types. 

\para{Simulator and Grasp Data.} 
We collected grasp data using NVIDIA Isaac Gym~\cite{makoviychuk2021isaac} which provides a highly parallelizable simulation of realistic grasps for various objects. Our simulation environment consists of a free-floating Panda gripper and an object mesh with no gravity. We used ShapeNet~\cite{chang2015shapenet} and 3DNet~\cite{wohlkinger20123dnet} objects from 10 categories, with 20 objects in each category (200 unique objects). The categories are mugs, hammer, bottles, boxes, cylinders, scissors, spatula, fork, pans and bowls. First, we sampled candidate grasps using the object's shape geometry as described in~\cite{mousavian20196} and randomly sampling within a bounding box that fully covers the object's shape. Any candidate in collision with the object was labelled as negative. The candidate was also negative if the closing volume between robot fingers had zero intersection with the object. Any remaining candidates were labelled via simulation using the Panda gripper moving through a predefined set of motions (forward/backward movements by 10 cm and 30 degree rotations around pitch axis). Positive grasps were those that successfully held on to the object after these motions. Using this methodology, we initially collected 12.5 million grasps (9.1\% positive labels).

\para{Improving the Discriminator.} Unfortunately, we found the standard classifier trained with the collected data above was unable to refine grasps well, particularly when the grasps were too far or too close to the object. Fortunately, the gradient flow formulation provides guidance into potential causes: refinement can fail when the classifier does not accurately reflect the density ratio and the resultant derivatives. This is likely to happen with learned classifiers in regions with low-data (in either class) where overfitting can occur. 

To mitigate this problem, we train our stability classifier with data augmentation (with random rotations and translations of the entire point cloud data) to ``fill in'' gaps in the data. Note that this data augmentation is performed in addition to the negative mining suggested in prior work~\cite{mousavian20196}. We also conducted \emph{positive} mining by perturbing good samples to ensure a sufficient number of positive samples. In total, we trained the model with 31.6 million grasps, which included 8.3 million (26\%) positive grasps. In addition, we found it important to include an auxiliary grasp pose reconstruction loss. 
This loss encouraged the top-level latent representation of the classifier to encode information about the grasp, which we believe facilitates generalization and better gradients; the classifier is given grasp information in a point cloud representation and forcing the network to recover the grasp pose mitigates over-fitting to the noisy object point cloud. 

\begin{figure}
    \centering
	\includegraphics[width=\columnwidth]{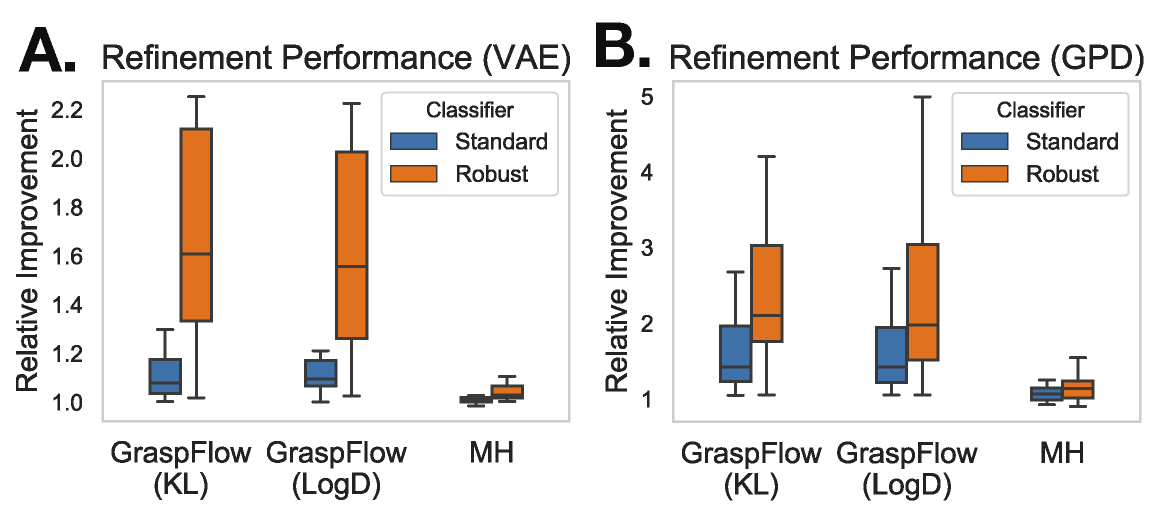}
	\caption{\small Refinement Performance for Stability evaluated using NVIDIA Isaac Gym.}
	\label{fig:simrefine}
\end{figure}

\para{Simulation Results.} With these changes, our \emph{robust} classifier was significantly better at refinement than the standard model. Fig. \ref{fig:simrefine}.A summarizes the relative improvement after refinement for 20 unseen ShapeNet objects. Here, the relative improvement captures the ratio of stable grasps before and after refinement per object (hence, values above 1 indicate a higher number of stable grasps). The initial 5000 grasps (per object) were sampled using the GraspNet VAE and stability was evaluated using  NVIDIA Isaac Gym. We refined grasps using \method{} with the KL and logD divergences (50 iterations), and a Metropolis-Hastings (MH) method (135 iterations to match computational time)~\cite{murali20206,mousavian20196}. 

The boxplots show that the refinement with the robust classifier resulted in more stable grasps compared to the standard classifier. In addition, prior work had noted a simple MH approach obtained similar performance to refinement via gradients~\cite{murali20206}. However, we observe the opposite: \method{} was better able to improve grasps compared to MH given the same computational budget. This discrepancy can potentially be explained by considering the importance of a robust classifier for refinement. 

\method{} can refine grasps from alternative samplers such as GPD~\cite{ten2017grasp}. \rev{Unlike the VAE, the GPD sampler uses heuristics to generate antipodal grasps.} Fig. \ref{fig:simrefine}.B again shows better refinement when using the robust classifier. In fact, the relative improvement here is larger as the GPD grasps were poorer for certain object classes; in total, the average proportion of stable grasps generated (before refinement) by the GraspNet VAE was 21\% (SD=11.3) compared to 14.7\% (SD=15.4) for GPD.

\subsection{Execution Classifier}
Unlike the stability classifier above, our execution classifier is theory-driven; it was developed using knowledge of robot kinematics. Here, our goal is to obtain grasps that avoid singular configurations of the robot. We will leverage manipulability ellipsoids, which are well-studied within the control community~\cite{yoshikawa1985manipulability, haviland2020purely}. 
For an open kinematic chain robot, the volume of the manipulability ellipsoid~\cite{lynch2017modern} is defined as:
\begin{equation} \label{eq:vol_man_ellipsoid}
    \omega(\btheta) = \sqrt{\det{\mathbf{J(\btheta)J(\btheta)^T}}} \geq 0
\end{equation}
where $\btheta$ is robot's joint configuration and $\mathbf{J(\btheta)}$ is Jacobian matrix. The volume $\omega(\mathbf{\btheta})$ spans the Cartesian speed of robot's end-effector and thus, it indicates singular or near-singular configurations of the robot~\cite{rubagotti2019semi}. As the eigenvalues of the manipulability ellipsoid are the reciprocal of the eigenvalues of the force ellipsoid, the volume also suggests how much force can be exerted on robot's end-effector. As such, it is desirable for the robot to operate in the configuration space that has a sufficiently large $\omega(\btheta)$. Our goal is to find a robot's joint configuration that ensures a minimum $\omega_{\text{th}}$, i.e.,  $\omega(\btheta) \geq \omega_{\text{th}}$. Using (\ref{eq:vol_man_ellipsoid}), we can define a classifier that gives the likelihood of executable grasp given robot's joint configuration, $\theta$:
\begin{equation}\label{eq:execution_classifier}
p(c_e=1|{\btheta}) = \sigma(C(\omega({\btheta}) - \omega_{\text{th}}))
\end{equation}
where $\sigma(\cdot)$ is the logistic function and $C$ is a scale coefficient. For the Panda robot, we set $C=100$ and $\omega_{\text{th}}=0.04$. Since our grasp $\grasp$ is represented in Cartesian space coordinates, we obtain the joint space configuration $\btheta$ using an \rev{analytical} inverse kinematics (IK) solver~\cite{he2021analytical}.

\rev{Grasps that were not reachable according to the IK solver were mapped to the closest reachable grasp in the joint space.}
We compute the Jacobian matrix in Eqn. (\ref{eq:vol_man_ellipsoid}) using the Newton-Euler method~\cite{sutanto2020encoding}.

\subsection{Handover Classifier}
We developed a simple handover classifier that leverages point-wise part segmentation labels using a pre-trained deep model~\cite{wang2019dynamic}. 
We then applied expert knowledge to select appropriate clusters to grasp for handover. 
Similar to the execution classifier, we compute likelihood of a positive handover grasp via a logistic classifier, $p(c_{\text{h}} = 1| \mathbf{g}) = \sigma (C\left((\mathbf{p}_{\text{centroid}} - \textbf{g}_{\text{translation}})^2-p_{\text{th}} \right))$ where $p_{\text{centroid}}$ is a centroid of the target cluster, $\textbf{g}_{\text{translation}}$ is translation of the grasp and  $C$ is a scale parameter and $p_{\text{th}}$ is the minimum set distance to the grasp. 
We set $p_{\text{th}} = 4$ cm and $C=10$ in our experiments.

%% file: content/experiments.tex
\begin{figure*}
\centering
	\includegraphics[width=.75\textwidth]{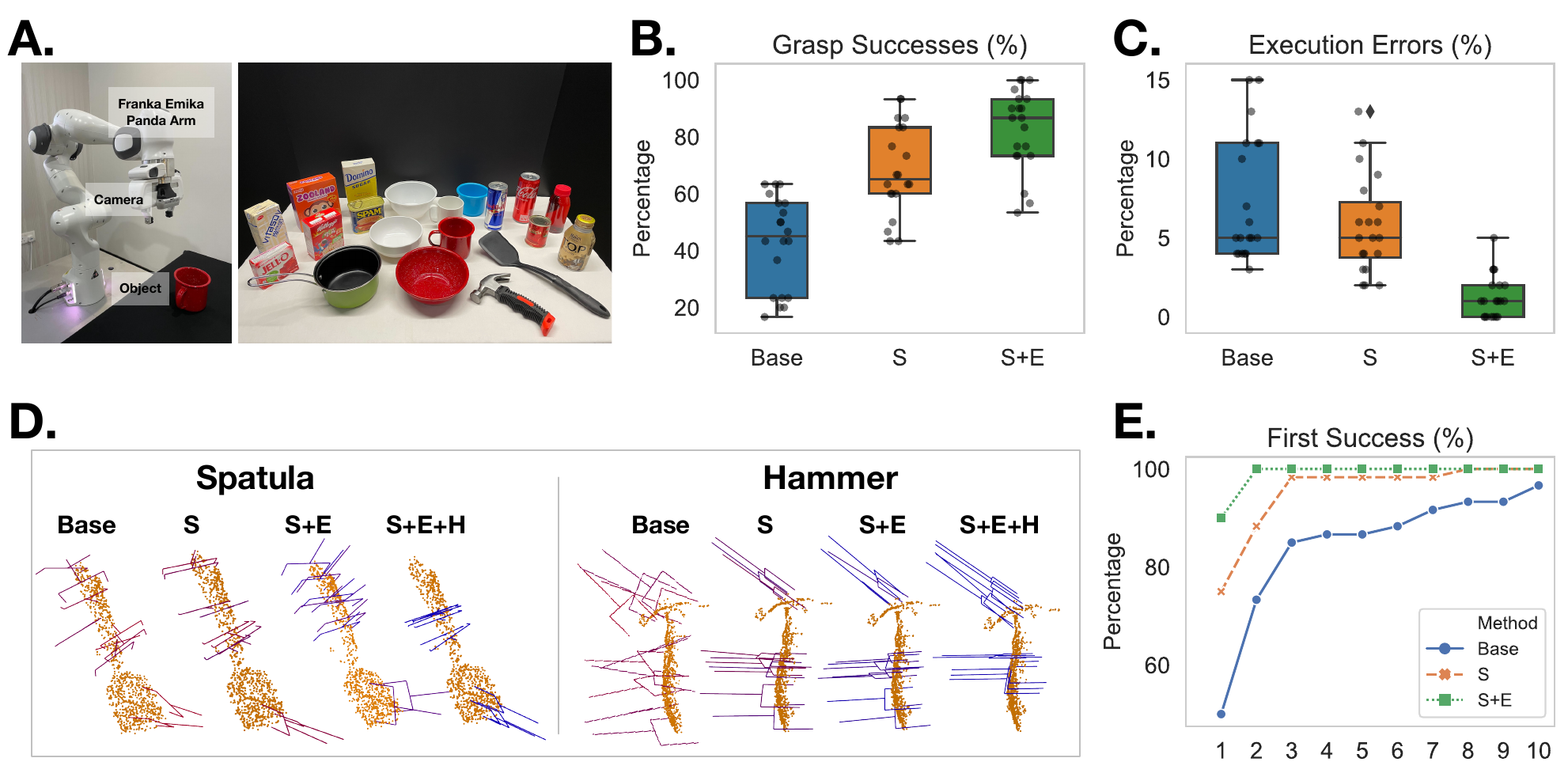}
	\caption{\small (\textbf{A}) Panda robot along with the objects used for grasping. (\textbf{B}) Percentage of successful grasp (stable and executable grasps). (\textbf{C}) Percentage of robot execution errors. (\textbf{D}) Example refined grasps using the different classifiers. (\textbf{E}) Number of grasps until the first successful grasp. }
	\label{fig:realworldresults}
\end{figure*}

\section{Real-World Experiments}

In this section, we report on experiments designed to validate our main claim, i.e., that \graspflow{} synthesizes successful grasps for real robots. We focused on obtaining stable and executable grasps for a 7-DoF Franka-Emika Panda robot equipped with an Intel RealSense RGB-D camera mounted on the arm, and include preliminary findings on obtaining functional grasps. 
Our experimental setup is shown in Fig. \ref{fig:realworldresults}.A and was constructed using ROS~\cite{quigley2009ros}; our code is available at \url{https://github.com/clear-nus/graspflow}. We used 20 unique household objects obtained from the YCB dataset~\cite{calli2017yale} or from a local supermarket. The objects are from known categories but were previously \emph{unseen}.

\para{Procedure.} First, an object was placed within the workspace of the robot. 
Then, the robot moves to 4 different poses around the object to collect point clouds using the RGB-D camera. At each pose, it stops for 5 seconds to minimize the effect of the noise. We filter out the background point cloud using~\cite{rusu20113d} and collect only the point cloud for the object of interest. Then, we map these point clouds into the world frame and combine them into a single dense point cloud. 

\para{Grasp Generation.} Given an object's point cloud, we use the GraspNet VAE to generate 200 grasp candidates. These candidates are evaluated using our stability classifier and we selected the top-10 highest scoring grasps; these grasps constitute our ``Base'' samples. We then applied \graspflow{} either with the stability classifier (S) or a combination of the stability and execution classifiers (S+E) to the Base samples. \graspflow{} parameters were set as $T=50$, $\eta_{\text{trans}}=10^{-5}$, $\eta_{\text{euler}}=10^{-4}$, and  $\gamma=10^{-4}$ across all objects. We repeated the same procedure three times for each object; in each one of these ``trials'', the object was placed in a random pose.  


\para{Grasp Evaluation.} In total, our experiment comprised 60 trials (20 objects, 3 trials each) and in each trial, we executed 10 grasps for each method (Base, S, S+E). In total, the robot executed 1800 grasps. 
\rev{We use Moveit!~\cite{chitta2012moveit} with the RRT connect planner~\cite{kuffner2000rrt} to plan a trajectory to the grasp.} 
A grasp was considered successful if the robot managed to grasp the object, lift it 20cm upwards, and hold it for 2 seconds. Otherwise, the grasp was labelled as a failure. We further distinguished failures as either a grasp failure or a robot execution failure (an error due to singularities). 


\subsection{Results}


\para{Does \method{} refinement improve grasp stability and execution?} In short, yes. The boxplot in Fig. \ref{fig:realworldresults}.B shows that the proportion of successful grasps (averaged across 20 different objects with 30 grasps each) increased from 42.8\% (SD=15.9) to 68.2\% (SD=15.5) when the stability (S) refinement was applied. The success percentage further increased to 82\% (SD=14.0) when both the stability and execution classifiers were used (S+E). These differences are statistically significant (assessed via three paired $t$-tests with Bonferroni-adjusted $\alpha=0.0033$ per test, $p$-value $< 10^{-5}$ across the pairwise differences). Likewise, Fig. \ref{fig:realworldresults}.C shows the percentage of robot errors experienced decreased sharply from 7.35\% (SD=3.9) to 1.3\% (SD=1.3) when the execution classifier was used with the stability classifier ($t(29) = 7.06$, $p<10^{-5}$). In contrast, refinement with the stability classifier only had a mild effect on error reduction; this result supports the need to refine using multiple criteria for the given context.

\begin{table}
\centering
    \caption{Handover Grasp Success}
    \begin{tabular}{l | c c }
        \hline 
        \textbf{Object} & \textbf{S+E} & \textbf{S+E+H} \\ \hline
        Hammer & 36.6\% & 60\% \\ 
        Spatula & 43.3\% & 76.7\% \\ \hline
    \end{tabular}
    \label{tbl:handover}
\end{table}
\para{Does incorporating the Handover discriminator enable functional grasping?} To answer this question, we conducted an experiment using the hammer and spatula objects to determine if \graspflow{} was able to generate grasps that were not only stable and executable, but also  functional. We used the same experimental protocol, with the additional success criterion that the executed grasp should be suitable for handover (ascertained by a human participant). The results are summarized in the Table \ref{tbl:handover} where (S+E+H) represents \graspflow{} with all three classifiers. We observe that incorporating the handover classifier increased the percentage of functional grasps by up to 33\%. 

\para{How quickly can we obtain a successful grasp?}
Fig.\ref{fig:realworldresults}.E shows the percentage of trials (out of 60) when the first successful grasp was obtained. Refinement with stability and execution classifiers lead to the first grasp being successful 94\% of the time and grasping was 100\% successful by the second grasp. For the handover experiment, 100\% grasp success  (i.e., stable, executable, and functional) was achieved by the second grasp.

%% file: content/discussion.tex
\section{Conclusions and Future Work}

This work presents \graspflow{}, an alternative form of grasp synthesis where grasps are refined/evolved to satisfy multiple criteria via differentiable discriminators. As experiments show, \graspflow{} generates more grasps candidates that are stable, executable, and functional compared to the baseline methods.

\para{Limitations and Future Work.} There are a number of ways that \method{} could be further improved. \rev{While our results are positive, it remains unclear how robust \method{} is to noise in the discriminator gradients.} Here, we used three classifiers and how the methodology scales to a large number of discriminators is an open question; the criteria landscape may be highly nonlinear with multiple minima, which can hamper sample evolution. The modularity afforded by the conditional independence assumption improves scaling in a computational sense, but the assumption may not hold with a large number of criteria. One possible workaround is to combine classifiers (with extra training) when the labels may be conditionally dependent. Further experiments are needed to examine such a setup. 
Finally, there are other contexts that are ripe for exploration---future work can look into \rev{applying \method{} with other methods such as DexNet}, functional grasping beyond handover, and also other contexts, e.g., grasping in clutter, with multi-fingered end-effectors, or for in-hand manipulation. 